\documentclass[11pt,a4paper]{article}
\usepackage[hyperref]{acl2019}
\usepackage{array}
\usepackage{times}
\usepackage{latexsym}
\usepackage{lscape}
\usepackage{booktabs}
\usepackage{longtable}
\usepackage{ngerman} 
\usepackage{url}
\usepackage{setspace}
\usepackage{hanging}
\usepackage{amsmath}
\usepackage{graphicx}
\usepackage{color, colortbl}

\definecolor{Gray}{gray}{0.84}
\aclfinalcopy 

\title{Linguistic evaluation of German-English Machine Translation using a Test Suite}

\author{Eleftherios Avramidis, Vivien Macketanz, Ursula Strohriegel, Hans
Uszkoreit \\
  German Research Center for Artificial Intelligence (DFKI), Berlin, Germany \\
  {\tt firstname.lastname@dfki.de} \\}

\date{}

\begin{document}
\maketitle
\begin{abstract}
We present the results of the application of a grammatical test suite 
for German$\rightarrow$English MT on the systems submitted at WMT19, 
with a detailed analysis for 107 phenomena organized in 14 categories. 
The systems still translate wrong one out of four test items in average. 
Low performance is indicated for idioms, modals, pseudo-clefts, multi-word 
expressions and verb valency.
When compared to last year, there has been a improvement of function words, non
verbal agreement and punctuation. More detailed conclusions about particular
systems and phenomena are also presented.

\end{abstract}

\section{Introduction}

For decades, the development of Machine Translation (MT) has been based on
either automatic metrics or human evaluation campaigns with the main focus on
producing scores or comparisons (rankings) expressing a generic notion of
quality.
Through the years there have been few examples of more detailed analyses of the
translation quality, both automatic (HTER
\cite{Snover:2009:FAH:1626431.1626480}, Hjerson \cite{popovic11:hjerson})  and
human \cite[MQM][]{pub7426}.
Nevertheless, these efforts have not been systematic and they have only focused
on few shallow error categories (e.g. morphology, lexical choice, reordering),
whereas the human evaluation campaigns have been limited by the requirement for
manual human effort.
Additionally, previous work on MT evaluation focused mostly on the ability of
the systems to translate test sets sampled from generic text sources, based on
the assumption that this text is representative of a common translation task
\cite{callisonburch-EtAl:2007:WMT}.

In order to provide more systematic methods to evaluate MT in a more
fine-grained level, recent research has relied to the idea of test suites
\cite{Guillou2016,Isabelle2017}.
The test suites are assembled in a way that allows testing particular issues
which are the focus of the evaluation.
The evaluation of the systems is not based on generic text samples,
but from the perspective of fulfilling a priori quality requirements.

In this paper we use the DFKI test suite for German$\rightarrow$English MT
\cite{Burchardt2017} in order to analyze the performance of the 16 MT Systems
that took part at the translation task of the Fourth Conference of Machine
Translation.
The evaluation focuses on 107 mostly grammatical phenomena organized in 14
categories.
In order to apply the test suite, we follow a semi-automatic methodology that
benefits from regular expressions, followed by minimal human refinement
(Section~\ref{sec:method}).
The application of the suite allows us to form conclusions on the particular
grammatical performance of the systems and perform several comparisons
(Section~\ref{sec:results}).

\section{Related Work}
Several test suites have been presented as part of the Test Suite track of the
Third Conference of Machine Translation \cite{WMT:2018}.
Each test suite focused on a particular phenomenon, such as discourse
\cite{bojar-EtAl:2018:WMT2}, morphology \cite{burlot-EtAl:2018:WMT}, grammatical
contrasts \cite{cinkova-bojar:2018:WMT}, pronouns \cite{guillou-EtAl:2018:WMT}
and word sense disambiguation \cite{rios-mller-sennrich:2018:WMT}.
In contrast to the above test suites, our test suite is the only one that does
such a systematic evaluation of more than one hundred phenomena.
A direct comparison can be done with the latter related paper, since it focuses
at the same language direction.
Its authors use automated methods to extract text items, whereas in our test
suite the test items are created manually.

\section{Method}
\label{sec:method}

The test suite is a manually devised test set whose contents are chosen with the
purpose to test the performance of the MT system on specific phenomena or
requirements related to quality.
For each phenomenon a subset of relevant test sentences is chosen manually.
Then, each MT system is requested to translate the given subset and the
performance of the system on the particular phenomenon is calculated based on
the percentage of the phenomenon instances that have been properly translated.

For this paper we use the latest version of the DFKI Test Suite for MT on German
to English.
The test suite has been presented in \cite{Burchardt2017} and applied
extensively in last year's shared task \cite{Macketanz2018}.
The current version contains 5560 test sentences in order to control 107
phenomena organised in 14 categories.
It is similar to the method used last year, with few minor corrections.
The number of the test instances per phenomenon varies, ranging between a 20 and
180 sentences.
A full list of the phenomena and their categories can be seen as part of the
results in the Appendix.
An example list of test sentences with correct and incorrect  translations is
available on GitHub\footnote{\url{https://github.com/DFKI-NLP/TQ_AutoTest}}.

\subsection{Construction and application of the test suite}

\begin{figure*}[ht]
\includegraphics[width=\textwidth]{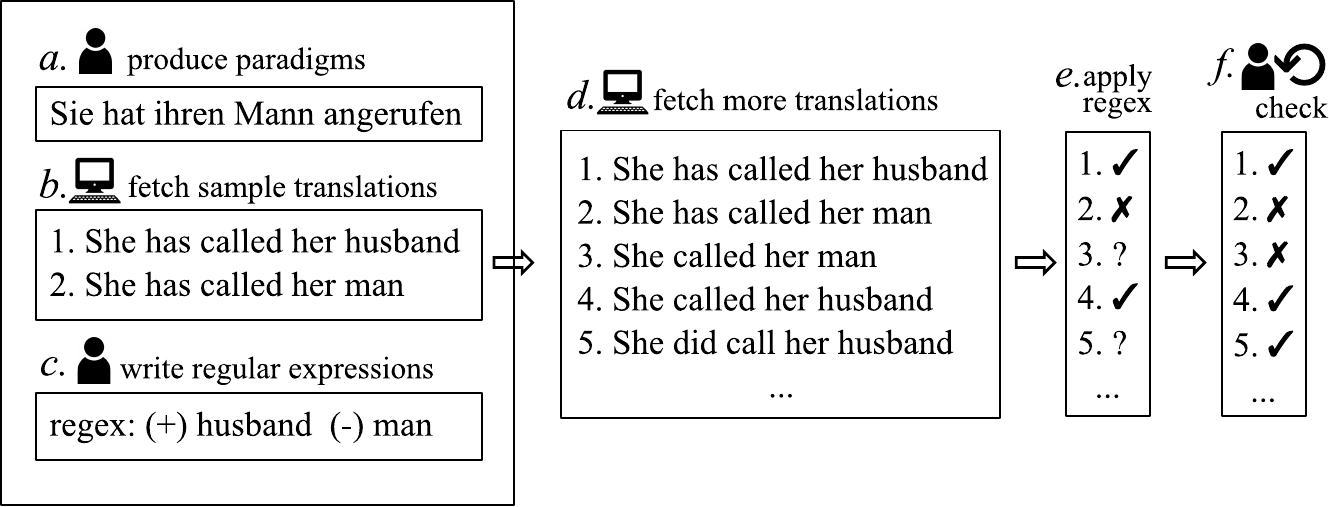}
\caption{Example of the preparation and application of the test suite for one
test sentence}
\label{fig:example}
\end{figure*}

The construction and the application of the test suite follows the steps below,
also indicated in Figure~\ref{fig:example}:

\noindent\textbf{(a) Produce paradigms}: 
A person with good knowledge of German and English grammar devises or selects a set of source language sentences that may trigger translation errors related to particular phenomena. 
These sentences may be written from scratch, inspired from previous observations on common MT errors or drawn from existing resources \cite{Lehmann1996}.

\noindent\textbf{(b) Fetch sample translations}:
The source sentences are given as an input to easily accessible MT systems and their outputs are fetched. 

\noindent\textbf{(c) Write regular expressions}:
By inspecting the MT output for every given sentence, the annotator writes rules that control whether the output contains a correct translation regarding the respective phenomenon. 
The rules are written as positive or negative regular expressions, that signify a correct or an incorrect translation respectively.

\noindent\textbf{(d) Fetch more translations}:
When the test suite contains a sufficient number of test items with the respective control rules, the test suite is ready for its broad application. 
The test items are consequently given to a large number of MT systems.
This is done in contact with their developers or through the submission process of a shared task, as is the case described in this paper.

\noindent\textbf{(e) Apply regular expressions}:
The control rules are applied on the MT outputs in order to check whether the
relevant phenomena have been translated properly.
When the MT output matches a positive regular expression, the translation is
considered correct (\textit{pass}) whereas when the MT output matches a negative
regular expression, the translation is considered incorrect (\textit{fail}).
Examples can be seen in Table~\ref{tab:examples}.

In case an MT output does not match either a positive or a negative regular
expression, or in case these contradict to each other, the automatic evaluation
results in a uncertain decision (\textit{warning}).

\begin{table}
\small
\begin{tabular}{ll}
\toprule
\rowcolor{Gray}
Lexical Ambiguity 	& \\
Das Gericht gestern Abend war lecker.	&\\
The court last night was delicious.	&fail\\
The dish last night was delicious.	&pass\\
\rowcolor{Gray}
Conditional	& \\
Er w\"urde einkaufen gehen, wenn die Gesch\"afte &\\
 nicht geschlossen h\"atten.	&\\
	He would go shopping if the stores didn't close.	&fail\\
	He would go shopping if the shops hadn't closed.	&pass\\
\rowcolor{Gray}
Passive voice	& \\
    Es wurde viel gefeiert und getanzt.	&\\
	A lot was celebrated and danced.	&fail\\
	There was a lot of celebration and dancing.	&pass\\
\bottomrule
\end{tabular}
\caption{Examples of passing and failing MT outputs}
\label{tab:examples}
\end{table}

\noindent\textbf{(f) Resolve warnings and refine regular expressions}:
The \textit{warnings} are given to the annotator, so that they manually resolve
them and if possible refine the rules to address similar cases in the future.
Through the iterative execution of steps (e) and (f) (which are an extension of
steps (c) and (d) respectively) the rules get more robust and attain a better
coverage.
Additionally, the annotator can add full sentences as rules, instead of regular
expressions, when

For every system we calculate the phenomenon-specific translation accuracy as the the number of the test sentences for the phenomenon which were translated properly, divided by the number of all test sentences for this phenomenon: 

\begin{equation*} 
 \textrm{accuracy} =  \frac{\textrm{correct\;translations}}{\textrm{sum\;of\;test\;sentences}} 
\end{equation*}

When doing comparisons, the significance of every comparison is confirmed with a
one-tailed Z-test with $\alpha = 0.95$.

\subsection{Experiment Setup}

In the evaluation presented in the paper, MT outputs are obtained from the 16
systems that are part of the \textit{news translation task} of the Fourth
Conference on Machine Translation (WMT19).
According to the details that the developers have published by the time this
paper is written, 10 of the systems are declared to be Neural Machine
Translation (NMT) systems and 9 of them confirm that they follow the Transformer
paradigm, whereas for the rest 6 systems no details were given.
For the evaluation of the MT outputs the software TQ-AutoTest
\cite{MACKETANZ18.121} was used.

After processing the MT output for the 5560 items of the test suite, the
automatic application of the regular expressions resulted to about 10\%
warnings.
Consequently, one human annotator (student of linguistics) committed about 70
hours of work in order to reduce the warnings to 3\%.
The final results were calculated using 5393 test items, which, after the manual
inspection, did not have any warning for any of the respective MT-outputs.

Since we applied the same test suite as last year, this year's automatic
evaluation is profiting from the manual refinement of the regular expressions
that took place last year.
The first application of the test suite in 2018 resulted in about 10-45\% of
warnings, whereas this year's application, we only had 8-28\%.
This year's results are therefore based on 16\% more valid test items, as
compared to last year.

\section{Results}
\label{sec:results}

The results of the test suite evaluation can be seen in
Tables~\ref{tab:categories} and \ref{tab:phenomena}, where the best systems for
every category or phenomenon are boldfaced.
The average accuracy per system is calculated either based on all test items
(with the assumption that all items have equal importance) or based on the
categories (with the assumption that all categories have equal importance).
In any case, since the averages are calculated on an artificial test suite and
not on a sample test set, one must be careful with their interpretation.

\subsection{Linguistic categories}

Despite the significant progress of NMT and the recent claims for human parity,
the results in terms of the test suite are somewhat mediocre.
The MT systems achieve 75.6\% accuracy in average for all given test items,
which indicates that one out of four test items is not translated properly.
If one considers the categories separately, only four categories have an
accuracy of more than 80\%: \textbf{negation}, where there are hardly any mistakes,
followed by \textbf{composition}, \textbf{function word} and \textbf{non-verbal
agreement}.
The lowest-performing categories are the \textbf{multi-word expressions} (MWE)
and the \textbf{verb valency} with about 66\% accuracy.

\subsection{Linguistic phenomena}
Most MT systems seem to struggle with \textbf{idioms}, since they could only
translate properly only 11.6\% of the ones in our test set, whereas a similar
situation can be observed with resultative predicates (17.8\%).
\textbf{Negated modal pluperfect} and \textbf{modal pluperfect} have an accuracy
of only 23-28\% and \textbf{pseudo-cleft} sentences of 36.6\%.
Some of the phenomena have an accuracy of about 50\%, in particular the
domain-specific terms, the pseudo-cleft clauses and the modal of pluperfect
subjunctive II (negated or not).
We may assume that these phenomena are not correctly translated because they do
not occur often enough in the training and development corpora.

On the other side, for quite a few phenomena an accuracy of more than 90\% has
been achieved.
This includes several cases of verbs declination concerning the transitive,
intransitive and ditransitive verbs mostly on perfect and future tenses, the
passive voice, the polar question, the infinitive clause, the conditional, the
focus particles, the location and the phrasal verbs.

\subsection{Comparison between systems}

As seen in Table~\ref{tab:categories}, the system that significantly wins most
categories is Facebook with 11 categories and an average of 87.5\% (if all
categories counted equally), followed by DFKI and RWTH which are in the best
cluster for 10 categories.
When it comes to averaging all test items, the best systems are RWTH and
online-A.
On specific categories, the most clear results come in \textbf{punctuation}
where NEU has the best performance with 100\% accuracy, whereas Online-X has the
worst with 31.7\%.
Concerning \textbf{ambiguity}, Facebook has the highest performance with 92.6\%
accuracy.
In \textbf{verb tense/aspect/mood}, RWTH Aachen and Online-A have the highest
performance with 84\% accuracy, whereas in this category, MSRA.MADL has the
lowest performance with 60.4\%.
For the rest of the categories there are small differences between the systems,
since more than five systems fall into the same significance cluster of the best
performance.

When looking into particular phenomena (Table~\ref{tab:phenomena}), Facebook has
the higher accuracy concerning \textbf{lexical ambiguity} with an accuracy of
93.7\%.
NEU and MSRA.MADL do best with more than 95\% on \textbf{quotation marks}.

The best system for translating \textbf{modal pluferect} is online-A with
75.6\%, whereas at the same category, online-Y and online-G perform worse with
less than 2.2\%.
On \textbf{modal negated - preterite}, the best systems are RWTH and UCAM with
more than 95\%.
On the contrary, MSRA.MADL achieves the worst accuracy, as compared to other
systems, in phenomena related to modals (perfect, present, preterite, negated
modal Future I), where it mistranslates half of the test items.
One system, Online-X, was the worst on quotation marks, as it did not convey
properly any of them, compared to other systems that did relatively well.
Online-Y also performs significantly worse than the other systems on
domain-specific terms. 

\subsection{Comparison with last year's systems}

One can attempt to do a vague comparison of the statistics between two
consequent years (Table~\ref{tab:yearscomparison}).
Here, the last column indicates the percentage of improvement from the average
accuracy of all systems from last year's shared task\footnote{unsupervised
systems excluded} to the average accuracy of all systems of this year.
Although this is not entirely accurate, since different systems participate, we
assume that the large amount of the test items allows some generalisations to
this direction.
When one compares the overall accuracy, there has been an improvement of about
6\%.
When focusing on particular categories, the biggest improvements are seen at
function words (+12.5\%), non-verbal agreement (+9.7\%) and punctuation (+8\%).
The smallest improvement is seen at named entity and terminology (+0.3\%).

\begin{table*}
\footnotesize\setlength{\tabcolsep}{4pt}
\begin{tabular}{lrrrrrrrrrrr}
\toprule
category	&\#	&JHU	&MLLP	&onlA	&onlB	&onlG	&onlY	&RWTH	&UCAM	&UEDIN	&avg\\
\midrule
Ambiguity                     	&74	&-2.7	&21.6	&4.1	&0.0	&4.1	&10.8	&-1.3	&2.7	&12.1	&6.9\\
Composition                   	&42	&4.8	&0.0	&14.3	&0.0	&9.5	&2.4	&-2.4	&-4.7	&7.1	&5.2\\
Coordination and ellipsis       	&23	&8.7	&-4.4	&0.0	&0.0	&13.1	&0.0	&0.0	&-13.1	&0.0	&7.3\\
False friends                 	&34	&-3.0	&5.8	&0.0	&3.0	&-5.9	&23.6	&5.9	&-5.8	&14.7	&6.8\\
Function word                 	&41	&-2.5	&7.3	&4.9	&0.0	&41.4	&0.0	&-7.4	&-2.4	&9.7	&12.5\\
LDD \& interrogatives          	&38	&10.6	&10.6	&-2.7	&0.0	&5.3	&0.0	&0.0	&5.3	&7.9	&5.6\\
MWE                           	&53	&5.6	&7.5	&5.7	&0.0	&1.9	&1.9	&3.8	&-1.8	&3.8	&4.7\\
Named entity and terminology   	&34	&5.9	&3.0	&5.9	&0.0	&-3.0	&-5.9	&8.9	&0.0	&5.9	&0.3\\
Negation                      	&19	&0.0	&0.0	&0.0	&0.0	&42.1	&0.0	&0.0	&0.0	&-10.5	&6.6\\
Non-verbal agreement          	&48	&12.5	&10.4	&12.5	&0.0	&22.9	&2.1	&-2.1	&0.0	&12.5	&9.7\\
Punctuation                   	&51	&5.9	&2.0	&-21.6	&0.0	&-7.9	&1.9	&27.5	&0.0	&23.5	&8.0\\
Subordination                 	&31	&3.3	&6.5	&-6.5	&3.2	&19.4	&3.2	&6.5	&0.0	&0.0	&5.0\\
Verb tense/aspect/mood        	&3995	&-4.0	&-5.9	&12.9	&0.2	&19.8	&1.6	&5.6	&-7.6	&5.1	&6.0\\
Verb valency                  	&30	&10.0	&0.0	&0.0	&0.0	&13.4	&6.6	&0.0	&0.0	&3.4	&5.8\\
\midrule
average (items)                	&4513	&-3.1	&-4.3	&11.6	&0.2	&18.7	&2.0
&5.3 &-6.8	&5.4	&6.1\\
average (categories)             &      & 3.9 &	4.6 &	2.1	& 0.5	& 12.6 &	3.4 &
3.2 & -2.0 & 6.8 & 6.5 \\
\bottomrule
\end{tabular}
    \caption{Percentage (\%) of accuracy improvement or deterioration
    between WMT18 and WMT19 for all the systems submitted (averaged in last column) and
    the systems submitted with the same name}
    \label{tab:yearscomparison}

\end{table*}

We also attempt to perform comparisons of the systems which were submitted with
the same name both years.
Again, the comparison should be done under the consideration that the MT systems
are different in many aspects, which are not possible to consider at the time
this paper is written.
The highest improvement is shown by the system online-G, which has an average
accuracy improvement of 18.7\%, with most remarkable the one concerning
negation, function words and non-verbal agreement.
Online-A has also improved at composition, verb issues and non-verbal agreement
and RWTH and UEDIN at punctuation. 
On the contrary, we can notice that UCAM deteriorated its accuracy for several
categories, mostly for coordination and ellipsis (-13.1\%), verb issues (-7.6\%)
and composition (-4.7\%).
JHU and Online-G and RWTH show some deterioration for three categories each,
whereas Online-A seems to have worsened considerably regarding
punctuation (-21.6\%) and UEDIN regarding negation (-10.5\%).

\section{Conclusion and Further Work}
\label{sec:conclusion}

The application of the test suite results in a multitude of findings of minor or
major importance.
Despite the recent advances, state-of-the-art German$\rightarrow$English MT
still translates erroneously one out of four test items of our test suite,
indicating that there is still room for improvement.
For instance, one can note the low performance on MWE and verb valency, whereas
there are issues with idioms, modals and pseudo-clefts.
Function words, non verbal agreement and punctuation on the other side have
significantly improved.

One potential benefit of the test suite would be to investigate the implication
of particular development settings and design decisions on particular phenomena.
For some superficial issues, such as punctuation, this would be relatively easy,
as pre- and post-processing steps may be responsible.
But for more complex phenomena, further comparative analysis of settings is
needed.
Unfortunately, this was hard to achieve for this shared task due to the
heterogeneity of the systems, but also due to the fact that at the time this
paper was written, no exact details about the systems were known.
We aim at looking further on such an analysis in future steps.

\section*{Acknowledgments}

This research was supported by the German Federal Ministry of Education and
Research through the projects DEEPLEE (01IW17001) and BBDC2 (01IS18025E).

\bibliography{acl2019}
\bibliographystyle{acl_natbib}

\onecolumn
\appendix

\begin{landscape}
\section{Appendices}

{\footnotesize\setlength{\tabcolsep}{4pt}
\begin{longtable}[c]{lrrrrrrrrrrrrrrrrrr}

\toprule
	&\#	&DFKI	&FB	&JHU	&MMLP	&MSRA	&NEU	&onlA	&onlB	&onlG	&onlX	&onlY	&PROMT	&RWTH	&Tartu	&UCAM	&UEDIN	&avg\\
\midrule
Ambiguity                     &   81 &          70.4  &  \textbf{92.6} &          64.2  &          76.5  &          80.2  &          75.3  &          69.1  &          76.5  &          72.8  &          50.6  &          76.5  &          48.1  &          77.8  &          60.5  &          75.3  &          59.3  & 70.4 \\
Composition                   &   48 &  \textbf{93.8} &  \textbf{97.9} &          87.5  &          85.4  &          83.3  &          87.5  &  \textbf{93.8} &  \textbf{95.8} &          83.3  &          58.3  &  \textbf{93.8} &          81.2  &          85.4  &          81.2  &  \textbf{89.6} &          87.5  & 86.6 \\
Coordination \& ellipsis      &   74 &  \textbf{85.1} &  \textbf{89.2} &  \textbf{78.4} &  \textbf{85.1} &          75.7  &  \textbf{81.1} &  \textbf{85.1} &  \textbf{85.1} &          60.8  &  \textbf{79.7} &  \textbf{78.4} &          74.3  &  \textbf{86.5} &          68.9  &  \textbf{78.4} &  \textbf{81.1} & 79.6 \\
False friends                 &   36 &          72.2  &          75.0  &          55.6  &          63.9  &          63.9  &          55.6  &          72.2  &  \textbf{77.8} &          72.2  &          72.2  &  \textbf{91.7} &          72.2  &          72.2  &          55.6  &          58.3  &          66.7  & 68.6 \\
Function word                 &   60 &  \textbf{88.3} &  \textbf{91.7} &          78.3  &  \textbf{91.7} &  \textbf{83.3} &  \textbf{90.0} &  \textbf{88.3} &          80.0  &  \textbf{90.0} &          65.0  &  \textbf{88.3} &  \textbf{85.0} &  \textbf{83.3} &          76.7  &  \textbf{88.3} &  \textbf{88.3} & 84.8 \\
LDD \& interrogatives         &  160 &  \textbf{82.5} &  \textbf{85.0} &  \textbf{79.4} &  \textbf{82.5} &  \textbf{81.2} &  \textbf{81.2} &          73.1  &  \textbf{78.8} &          66.2  &          63.1  &          75.6  &          71.2  &  \textbf{83.8} &          76.2  &  \textbf{85.0} &          69.4  & 77.1 \\
MWE                           &   77 &  \textbf{68.8} &  \textbf{77.9} &  \textbf{64.9} &  \textbf{66.2} &  \textbf{66.2} &  \textbf{67.5} &  \textbf{67.5} &  \textbf{70.1} &  \textbf{68.8} &          48.1  &  \textbf{71.4} &          55.8  &  \textbf{70.1} &          61.0  &  \textbf{63.6} &          62.3  & 65.7 \\
Named entity \& terminology  &   87 &  \textbf{80.5} &  \textbf{82.8} &  \textbf{83.9} &  \textbf{81.6} &  \textbf{82.8} &  \textbf{79.3} &  \textbf{81.6} &  \textbf{85.1} &          66.7  &          48.3  &  \textbf{82.8} &          64.4  &  \textbf{85.1} &  \textbf{79.3} &  \textbf{80.5} &  \textbf{83.9} & 78.0 \\
Negation                      &   20 &         100.0  &         100.0  &         100.0  &         100.0  &          95.0  &         100.0  &         100.0  &          95.0  &         100.0  &         100.0  &         100.0  &          90.0  &         100.0  &         100.0  &         100.0  &          90.0  & 98.1 \\
Non-verbal agreement          &   61 &  \textbf{85.2} &  \textbf{91.8} &  \textbf{83.6} &  \textbf{88.5} &  \textbf{86.9} &          78.7  &  \textbf{83.6} &  \textbf{86.9} &          80.3  &          65.6  &          80.3  &          70.5  &  \textbf{82.0} &          80.3  &          78.7  &  \textbf{82.0} & 81.6 \\
Punctuation                   &   60 &          85.0  &          93.3  &          70.0  &          68.3  &          95.0  & \textbf{100.0} &          76.7  &          76.7  &          58.3  &          31.7  &          80.0  &          83.3  &          88.3  &          91.7  &          58.3  &          90.0  & 77.9 \\
Subordination                 &  168 &  \textbf{89.3} &  \textbf{89.9} &  \textbf{88.7} &  \textbf{89.9} &  \textbf{88.1} &  \textbf{85.7} &          75.6  &  \textbf{85.7} &  \textbf{83.3} &          70.8  &  \textbf{86.3} &          79.2  &  \textbf{88.7} &  \textbf{83.9} &  \textbf{89.9} &          76.2  & 84.4 \\
Verb tense/aspect/mood        & 4375 &          77.1  &          79.4  &          70.3  &          78.8  &          60.4  &          77.1  &  \textbf{84.1} &          74.3  &          66.2  &          70.2  &          72.7  &          75.4  &  \textbf{83.9} &          71.7  &          79.2  &          81.4  & 75.1 \\
Verb valency                  &   86 &  \textbf{72.1} &  \textbf{79.1} &  \textbf{68.6} &  \textbf{67.4} &  \textbf{70.9} &  \textbf{66.3} &  \textbf{67.4} &  \textbf{68.6} &  \textbf{67.4} &          55.8  &  \textbf{66.3} &          54.7  &  \textbf{72.1} &          62.8  &  \textbf{68.6} &          60.5  & 66.8 \\
\midrule
average (items)              & 5393 &          78.0  &          80.9  & 
71.6  &          79.2  &          64.3  &          77.7  &  \textbf{82.8} &          75.5  &          67.5  &          68.4  &          74.1  &          74.4  &  \textbf{83.6} &          72.3  &          79.2  &          80.2  & 75.6 \\
average (categories)          & 5393 &          82.2  &  \textbf{87.5} &         
76.7 &          80.4  &          79.5  &          80.4  &          79.9  &          81.2  &          74.0  &          62.8  &          81.7  &          71.8  &          82.8  &          75.0  &          78.1  &          77.0  & 78.2 \\
\bottomrule
   
      \caption{Accuracies of successful translations for 16 systems and 14
      categories.
      Boldface indicates significantly best systems in each row}
\label{tab:categories}
\end{longtable}
}
{\footnotesize\setlength{\tabcolsep}{4pt}

\begin{longtable}[c]{lrrrrrrrrrrrrrrrrrr}

    \toprule
    	&\#	&DFKI	&FB	&JHU	&MLLP	&MSRA	&NEU	&onlA	&onlB	&onlG	&onlX	&onlY	&PROMT	&RWTH	&Tartu	&UCAM	&UEDIN &avg\\
    \midrule
    \endfirsthead
    \toprule
    	&\#	&DFKI	&FB	&JHU	&MLLP	&MSRA	&NEU	&onlA	&onlB	&onlG	&onlX	&onlY	&PROMT	&RWTH	&Tartu	&UCAM	&UEDIN &avg\\
    \midrule
    \endhead
    \bottomrule
    \endfoot
    \endlastfoot
\rowcolor{Gray}
Ambiguity                     &   81 &          70.4  &  \textbf{92.6} &          64.2  &          76.5  &          80.2  &          75.3  &          69.1  &          76.5  &          72.8  &          50.6  &          76.5  &          48.1  &          77.8  &          60.5  &          75.3  &          59.3  & 70.4 \\
Lexical ambiguity             &   63 &          73.0  &  \textbf{93.7} &          65.1  &          77.8  &          81.0  &          74.6  &          73.0  &          82.5  &          79.4  &          55.6  &          82.5  &          50.8  &          81.0  &          58.7  &          76.2  &          66.7  & 73.2 \\
Structural ambiguity          &   18 &  \textbf{61.1} &  \textbf{88.9} &  \textbf{61.1} &  \textbf{72.2} &  \textbf{77.8} &  \textbf{77.8} &          55.6  &          55.6  &          50.0  &          33.3  &          55.6  &          38.9  &  \textbf{66.7} &  \textbf{66.7} &  \textbf{72.2} &          33.3  & 60.4 \\
\rowcolor{Gray}
Composition                   &   48 &  \textbf{93.8} &  \textbf{97.9} &          87.5  &          85.4  &          83.3  &          87.5  &  \textbf{93.8} &  \textbf{95.8} &          83.3  &          58.3  &  \textbf{93.8} &          81.2  &          85.4  &          81.2  &  \textbf{89.6} &          87.5  & 86.6 \\
Compound                      &   28 &  \textbf{92.9} &  \textbf{96.4} &  \textbf{82.1} &          78.6  &          78.6  &  \textbf{82.1} &  \textbf{92.9} &  \textbf{96.4} &  \textbf{82.1} &          50.0  &  \textbf{89.3} &          67.9  &  \textbf{82.1} &  \textbf{85.7} &  \textbf{92.9} &          78.6  & 83.0 \\
Phrasal verb                  &   20 &  \textbf{95.0} & \textbf{100.0} &  \textbf{95.0} &  \textbf{95.0} &  \textbf{90.0} &  \textbf{95.0} &  \textbf{95.0} &  \textbf{95.0} &          85.0  &          70.0  & \textbf{100.0} & \textbf{100.0} &  \textbf{90.0} &          75.0  &          85.0  & \textbf{100.0} & 91.6 \\
\rowcolor{Gray}
Coordination \& ellipsis      &   74 &  \textbf{85.1} &  \textbf{89.2} &  \textbf{78.4} &  \textbf{85.1} &          75.7  &  \textbf{81.1} &  \textbf{85.1} &  \textbf{85.1} &          60.8  &  \textbf{79.7} &  \textbf{78.4} &          74.3  &  \textbf{86.5} &          68.9  &  \textbf{78.4} &  \textbf{81.1} & 79.6 \\
Gapping                       &   19 &  \textbf{94.7} & \textbf{100.0} &  \textbf{94.7} & \textbf{100.0} & \textbf{100.0} &  \textbf{89.5} &  \textbf{89.5} &  \textbf{89.5} &          57.9  &  \textbf{89.5} &          73.7  &          73.7  &  \textbf{94.7} &          78.9  &  \textbf{94.7} &  \textbf{89.5} & 88.2 \\
Right node raising            &   20 &  \textbf{80.0} &  \textbf{85.0} &  \textbf{80.0} &  \textbf{75.0} &          55.0  &  \textbf{85.0} &  \textbf{85.0} &  \textbf{85.0} &          50.0  &  \textbf{70.0} &  \textbf{75.0} &  \textbf{70.0} &  \textbf{80.0} &  \textbf{60.0} &  \textbf{60.0} &  \textbf{60.0} & 72.2 \\
Sluicing                      &   18 &          88.9  &          88.9  &          83.3  &          88.9  &          88.9  &          88.9  &          88.9  &          88.9  &          83.3  &          77.8  &          88.9  &          83.3  &          88.9  &          88.9  &          88.9  &          88.9  & 87.2 \\
Stripping                     &   17 &  \textbf{76.5} &  \textbf{82.4} &          52.9  &  \textbf{76.5} &  \textbf{58.8} &  \textbf{58.8} &  \textbf{76.5} &  \textbf{76.5} &          52.9  &  \textbf{82.4} &  \textbf{76.5} &  \textbf{70.6} &  \textbf{82.4} &          47.1  &  \textbf{70.6} &  \textbf{88.2} & 70.6 \\
\rowcolor{Gray}
False friends                 &   36 &          72.2  &          75.0  &          55.6  &          63.9  &          63.9  &          55.6  &          72.2  &  \textbf{77.8} &          72.2  &          72.2  &  \textbf{91.7} &          72.2  &          72.2  &          55.6  &          58.3  &          66.7  & 68.6 \\
\rowcolor{Gray}
Function word                 &   60 &  \textbf{88.3} &  \textbf{91.7} &          78.3  &  \textbf{91.7} &  \textbf{83.3} &  \textbf{90.0} &  \textbf{88.3} &          80.0  &  \textbf{90.0} &          65.0  &  \textbf{88.3} &  \textbf{85.0} &  \textbf{83.3} &          76.7  &  \textbf{88.3} &  \textbf{88.3} & 84.8 \\
Focus particle                &   20 &  \textbf{95.0} & \textbf{100.0} &  \textbf{95.0} &  \textbf{90.0} & \textbf{100.0} &  \textbf{95.0} &          85.0  &  \textbf{95.0} &  \textbf{90.0} &          85.0  &  \textbf{95.0} &          85.0  &  \textbf{95.0} &  \textbf{95.0} &  \textbf{95.0} & \textbf{100.0} & 93.4 \\
Modal particle                &   22 &  \textbf{81.8} &  \textbf{81.8} &  \textbf{81.8} &  \textbf{86.4} &  \textbf{72.7} &  \textbf{81.8} &  \textbf{81.8} &  \textbf{77.3} &  \textbf{90.9} &          63.6  &  \textbf{81.8} &  \textbf{86.4} &          68.2  &  \textbf{77.3} &  \textbf{77.3} &          68.2  & 78.7 \\
Question tag                  &   18 &  \textbf{88.9} &  \textbf{94.4} &          55.6  & \textbf{100.0} &          77.8  &  \textbf{94.4} & \textbf{100.0} &          66.7  &  \textbf{88.9} &          44.4  &  \textbf{88.9} &          83.3  &  \textbf{88.9} &          55.6  &  \textbf{94.4} & \textbf{100.0} & 82.6 \\
\rowcolor{Gray}
LDD \& interrogatives         &  160 &  \textbf{82.5} &  \textbf{85.0} &  \textbf{79.4} &  \textbf{82.5} &  \textbf{81.2} &  \textbf{81.2} &          73.1  &  \textbf{78.8} &          66.2  &          63.1  &          75.6  &          71.2  &  \textbf{83.8} &          76.2  &  \textbf{85.0} &          69.4  & 77.1 \\
Extended adjective construction &   18 &  \textbf{83.3} &  \textbf{83.3} &  \textbf{66.7} &  \textbf{83.3} &  \textbf{61.1} &  \textbf{77.8} &  \textbf{66.7} &  \textbf{66.7} &          44.4  &          38.9  &  \textbf{66.7} &  \textbf{61.1} &  \textbf{83.3} &  \textbf{61.1} &  \textbf{72.2} &  \textbf{66.7} & 67.7 \\
Extraposition                 &   18 &          44.4  &          61.1  &          55.6  &          55.6  &          72.2  &          66.7  &          55.6  &          61.1  &          50.0  &          50.0  &          61.1  &          66.7  &          61.1  &          55.6  &          61.1  &          55.6  & 58.3 \\
Multiple connectors           &   20 &  \textbf{90.0} &  \textbf{85.0} &  \textbf{80.0} &  \textbf{80.0} &  \textbf{80.0} &  \textbf{85.0} &  \textbf{75.0} &  \textbf{75.0} &  \textbf{65.0} &  \textbf{80.0} &          55.0  &  \textbf{85.0} &  \textbf{85.0} &  \textbf{85.0} &  \textbf{85.0} &  \textbf{70.0} & 78.8 \\
Pied-piping                   &   19 &  \textbf{84.2} &  \textbf{84.2} &  \textbf{89.5} &  \textbf{78.9} &  \textbf{89.5} &  \textbf{84.2} &  \textbf{78.9} &  \textbf{73.7} &  \textbf{73.7} &          52.6  &  \textbf{84.2} &  \textbf{73.7} &  \textbf{84.2} &          68.4  &  \textbf{94.7} &  \textbf{73.7} & 79.3 \\
Polar question                &   19 & \textbf{100.0} & \textbf{100.0} & \textbf{100.0} & \textbf{100.0} & \textbf{100.0} &  \textbf{89.5} &          84.2  & \textbf{100.0} &          84.2  & \textbf{100.0} &  \textbf{94.7} &  \textbf{94.7} & \textbf{100.0} & \textbf{100.0} & \textbf{100.0} &          84.2  & 95.7 \\
Scrambling                    &   17 &  \textbf{76.5} &  \textbf{88.2} &  \textbf{70.6} &  \textbf{76.5} &  \textbf{76.5} &  \textbf{70.6} &          52.9  &  \textbf{70.6} &  \textbf{64.7} &          29.4  &  \textbf{70.6} &          29.4  &  \textbf{64.7} &  \textbf{70.6} &  \textbf{88.2} &          35.3  & 64.7 \\
Topicalization                &   18 &  \textbf{83.3} &  \textbf{83.3} &  \textbf{77.8} &  \textbf{83.3} &  \textbf{77.8} &  \textbf{72.2} &  \textbf{61.1} &  \textbf{77.8} &  \textbf{66.7} &  \textbf{66.7} &  \textbf{66.7} &  \textbf{61.1} &  \textbf{88.9} &  \textbf{66.7} &  \textbf{83.3} &          55.6  & 73.3 \\
Wh-movement                   &   31 &  \textbf{90.3} &  \textbf{90.3} &  \textbf{87.1} &  \textbf{93.5} &  \textbf{87.1} &  \textbf{93.5} &  \textbf{93.5} &  \textbf{93.5} &          74.2  &          74.2  &  \textbf{93.5} &  \textbf{83.9} &  \textbf{93.5} &  \textbf{90.3} &  \textbf{90.3} &  \textbf{93.5} & 88.9 \\
\rowcolor{Gray}
MWE                           &   77 &  \textbf{68.8} &  \textbf{77.9} &  \textbf{64.9} &  \textbf{66.2} &  \textbf{66.2} &  \textbf{67.5} &  \textbf{67.5} &  \textbf{70.1} &  \textbf{68.8} &          48.1  &  \textbf{71.4} &          55.8  &  \textbf{70.1} &          61.0  &  \textbf{63.6} &          62.3  & 65.7 \\
Collocation                   &   19 &          68.4  &  \textbf{94.7} &          57.9  &          68.4  &  \textbf{78.9} &  \textbf{73.7} &  \textbf{78.9} &  \textbf{84.2} &  \textbf{78.9} &          52.6  &  \textbf{89.5} &          57.9  &  \textbf{73.7} &          63.2  &          57.9  &          63.2  & 71.4 \\
Idiom                         &   20 &          15.0  &          20.0  &          15.0  &           5.0  &          15.0  &           5.0  &          15.0  &          15.0  &          10.0  &          10.0  &          10.0  &           5.0  &          20.0  &          10.0  &           5.0  &          10.0  & 11.6 \\
Prepositional MWE             &   19 & \textbf{100.0} & \textbf{100.0} & \textbf{100.0} &  \textbf{94.7} &  \textbf{89.5} & \textbf{100.0} &          84.2  &  \textbf{94.7} &  \textbf{94.7} &          57.9  &  \textbf{89.5} &          73.7  & \textbf{100.0} &          78.9  & \textbf{100.0} &  \textbf{89.5} & 90.5 \\
Verbal MWE                    &   19 &  \textbf{94.7} & \textbf{100.0} &  \textbf{89.5} & \textbf{100.0} &          84.2  &  \textbf{94.7} &  \textbf{94.7} &  \textbf{89.5} &  \textbf{94.7} &          73.7  & \textbf{100.0} &  \textbf{89.5} &  \textbf{89.5} &  \textbf{94.7} &  \textbf{94.7} &  \textbf{89.5} & 92.1 \\
\rowcolor{Gray}
Named entity \& terminology  &   87 &  \textbf{80.5} &  \textbf{82.8} &  \textbf{83.9} &  \textbf{81.6} &  \textbf{82.8} &  \textbf{79.3} &  \textbf{81.6} &  \textbf{85.1} &          66.7  &          48.3  &  \textbf{82.8} &          64.4  &  \textbf{85.1} &  \textbf{79.3} &  \textbf{80.5} &  \textbf{83.9} & 78.0 \\
Date                          &   20 &  \textbf{85.0} &  \textbf{90.0} &  \textbf{90.0} &  \textbf{95.0} &  \textbf{95.0} &  \textbf{90.0} &  \textbf{95.0} &  \textbf{95.0} &          50.0  &          55.0  &  \textbf{95.0} &          50.0  &  \textbf{90.0} &  \textbf{95.0} &  \textbf{95.0} &  \textbf{95.0} & 85.0 \\
Domainspecific term           &   19 &  \textbf{57.9} &  \textbf{68.4} &  \textbf{63.2} &  \textbf{52.6} &  \textbf{57.9} &  \textbf{52.6} &  \textbf{52.6} &  \textbf{68.4} &  \textbf{42.1} &          21.1  &  \textbf{47.4} &  \textbf{36.8} &  \textbf{68.4} &  \textbf{52.6} &  \textbf{57.9} &  \textbf{57.9} & 53.6 \\
Location                      &   20 &  \textbf{95.0} &  \textbf{95.0} & \textbf{100.0} &  \textbf{95.0} &  \textbf{90.0} &  \textbf{90.0} & \textbf{100.0} &  \textbf{90.0} &          80.0  &          65.0  &  \textbf{90.0} &  \textbf{90.0} &  \textbf{95.0} &  \textbf{90.0} &  \textbf{95.0} & \textbf{100.0} & 91.2 \\
Measuring unit                &   19 &          84.2  &          84.2  &  \textbf{94.7} &  \textbf{89.5} &  \textbf{89.5} &  \textbf{89.5} &  \textbf{89.5} &  \textbf{89.5} &  \textbf{89.5} &          63.2  & \textbf{100.0} &  \textbf{89.5} &  \textbf{94.7} &          78.9  &          78.9  & \textbf{100.0} & 87.8 \\
Proper name                   &    9 &          77.8  &          66.7  &          55.6  &          66.7  &          77.8  &          66.7  &          55.6  &          77.8  &          77.8  &          22.2  &          77.8  &          44.4  &          66.7  &          77.8  &          66.7  &          44.4  & 63.9 \\
\rowcolor{Gray}
Negation                      &   20 &         100.0  &         100.0  &         100.0  &         100.0  &          95.0  &         100.0  &         100.0  &          95.0  &         100.0  &         100.0  &         100.0  &          90.0  &         100.0  &         100.0  &         100.0  &          90.0  & 98.1 \\
\rowcolor{Gray}
Non-verbal agreement          &   61 &  \textbf{85.2} &  \textbf{91.8} &  \textbf{83.6} &  \textbf{88.5} &  \textbf{86.9} &          78.7  &  \textbf{83.6} &  \textbf{86.9} &          80.3  &          65.6  &          80.3  &          70.5  &  \textbf{82.0} &          80.3  &          78.7  &  \textbf{82.0} & 81.6 \\
Coreference                   &   20 &  \textbf{75.0} &  \textbf{85.0} &  \textbf{75.0} &  \textbf{80.0} &  \textbf{80.0} &  \textbf{65.0} &  \textbf{75.0} &  \textbf{70.0} &  \textbf{65.0} &          45.0  &  \textbf{65.0} &  \textbf{65.0} &  \textbf{65.0} &  \textbf{70.0} &  \textbf{70.0} &  \textbf{80.0} & 70.6 \\
External possessor            &   21 &  \textbf{85.7} &  \textbf{95.2} &  \textbf{76.2} &  \textbf{90.5} &  \textbf{81.0} &  \textbf{76.2} &  \textbf{81.0} &  \textbf{90.5} &  \textbf{81.0} &          61.9  &  \textbf{81.0} &          57.1  &  \textbf{85.7} &  \textbf{81.0} &          71.4  &          71.4  & 79.2 \\
Internal possessor            &   20 &          95.0  &          95.0  &         100.0  &          95.0  &         100.0  &          95.0  &          95.0  &         100.0  &          95.0  &          90.0  &          95.0  &          90.0  &          95.0  &          90.0  &          95.0  &          95.0  & 95.0 \\
\rowcolor{Gray}
Punctuation                   &   60 &          85.0  &          93.3  &          70.0  &          68.3  &          95.0  & \textbf{100.0} &          76.7  &          76.7  &          58.3  &          31.7  &          80.0  &          83.3  &          88.3  &          91.7  &          58.3  &          90.0  & 77.9 \\
Comma                         &   20 &         100.0  &         100.0  &         100.0  &          95.0  &          95.0  &         100.0  &         100.0  &          95.0  &          95.0  &          95.0  &         100.0  &         100.0  &         100.0  &          95.0  &         100.0  &          95.0  & 97.8 \\
Quotation marks               &   40 &          77.5  &          90.0  &          55.0  &          55.0  &  \textbf{95.0} & \textbf{100.0} &          65.0  &          67.5  &          40.0  &           0.0  &          70.0  &          75.0  &          82.5  &          90.0  &          37.5  &          87.5  & 68.0 \\
\rowcolor{Gray}
Subordination                 &  168 &  \textbf{89.3} &  \textbf{89.9} &  \textbf{88.7} &  \textbf{89.9} &  \textbf{88.1} &  \textbf{85.7} &          75.6  &  \textbf{85.7} &  \textbf{83.3} &          70.8  &  \textbf{86.3} &          79.2  &  \textbf{88.7} &  \textbf{83.9} &  \textbf{89.9} &          76.2  & 84.4 \\
Adverbial clause              &   20 &          90.0  &          90.0  &          95.0  &          90.0  &          95.0  &          90.0  &          85.0  &          90.0  &          90.0  &          75.0  &          95.0  &          90.0  &          95.0  &          80.0  &          90.0  &          85.0  & 89.1 \\
Cleft sentence                &   19 &  \textbf{94.7} &  \textbf{94.7} &  \textbf{94.7} &  \textbf{94.7} & \textbf{100.0} &  \textbf{94.7} &          84.2  &  \textbf{89.5} &          84.2  &          84.2  &          84.2  &          78.9  & \textbf{100.0} &  \textbf{94.7} & \textbf{100.0} &  \textbf{89.5} & 91.4 \\
Free relative clause          &   18 &          94.4  &          83.3  &          83.3  &          94.4  &          94.4  &          94.4  &          94.4  &          88.9  &          88.9  &          94.4  &          88.9  &          94.4  &          94.4  &          88.9  &          88.9  &          94.4  & 91.3 \\
Indirect speech               &   19 &  \textbf{73.7} &  \textbf{84.2} &  \textbf{89.5} &  \textbf{89.5} &  \textbf{73.7} &          68.4  &          42.1  &  \textbf{94.7} &  \textbf{84.2} &          57.9  &  \textbf{73.7} &          63.2  &  \textbf{78.9} &          42.1  &  \textbf{84.2} &          36.8  & 71.1 \\
Infinitive clause             &   20 & \textbf{100.0} & \textbf{100.0} &  \textbf{95.0} &  \textbf{90.0} & \textbf{100.0} &  \textbf{90.0} &          85.0  &  \textbf{95.0} &  \textbf{95.0} &  \textbf{90.0} & \textbf{100.0} &  \textbf{90.0} & \textbf{100.0} & \textbf{100.0} & \textbf{100.0} &          85.0  & 94.7 \\
Object clause                 &   20 &  \textbf{95.0} & \textbf{100.0} &  \textbf{90.0} &  \textbf{95.0} &  \textbf{95.0} &  \textbf{95.0} &          85.0  &  \textbf{95.0} &  \textbf{95.0} &          85.0  &  \textbf{90.0} &          85.0  &  \textbf{95.0} &          85.0  &  \textbf{95.0} &  \textbf{90.0} & 91.9 \\
Pseudo-cleft sentence         &   18 &  \textbf{66.7} &  \textbf{72.2} &  \textbf{66.7} &  \textbf{72.2} &  \textbf{61.1} &  \textbf{55.6} &          22.2  &  \textbf{50.0} &  \textbf{55.6} &          22.2  &  \textbf{61.1} &  \textbf{44.4} &  \textbf{55.6} &  \textbf{77.8} &  \textbf{61.1} &          22.2  & 54.2 \\
Relative clause               &   18 &          94.4  &          83.3  &          83.3  &          94.4  &          77.8  &          83.3  &          83.3  &          77.8  &          77.8  &          83.3  &          94.4  &          83.3  &          83.3  &          88.9  &          88.9  &          88.9  & 85.4 \\
Subject clause                &   16 &  \textbf{93.8} & \textbf{100.0} & \textbf{100.0} &  \textbf{87.5} &  \textbf{93.8} & \textbf{100.0} & \textbf{100.0} &  \textbf{87.5} &          75.0  &          37.5  &  \textbf{87.5} &          81.2  &  \textbf{93.8} & \textbf{100.0} & \textbf{100.0} &  \textbf{93.8} & 89.5 \\
\rowcolor{Gray}
Verb tense/aspect/mood        & 4375 &          77.1  &          79.4  &          70.3  &          78.8  &          60.4  &          77.1  &  \textbf{84.1} &          74.3  &          66.2  &          70.2  &          72.7  &          75.4  &  \textbf{83.9} &          71.7  &          79.2  &          81.4  & 75.1 \\
Conditional                   &   19 & \textbf{100.0} &  \textbf{89.5} &          84.2  & \textbf{100.0} &  \textbf{89.5} & \textbf{100.0} &  \textbf{94.7} & \textbf{100.0} &  \textbf{89.5} &          68.4  & \textbf{100.0} &  \textbf{94.7} & \textbf{100.0} &          84.2  & \textbf{100.0} & \textbf{100.0} & 93.4 \\
Ditransitive - future I       &   36 & \textbf{100.0} & \textbf{100.0} & \textbf{100.0} & \textbf{100.0} &  \textbf{97.2} &          83.3  & \textbf{100.0} & \textbf{100.0} &          91.7  & \textbf{100.0} & \textbf{100.0} & \textbf{100.0} & \textbf{100.0} &  \textbf{94.4} &          83.3  & \textbf{100.0} & 96.9 \\
Ditransitive - future I subjunctive II &   36 & \textbf{100.0} & \textbf{100.0} &          91.7  & \textbf{100.0} & \textbf{100.0} &          80.6  & \textbf{100.0} &  \textbf{97.2} &  \textbf{97.2} & \textbf{100.0} & \textbf{100.0} &  \textbf{97.2} & \textbf{100.0} &          88.9  &          83.3  &  \textbf{97.2} & 95.8 \\
Ditransitive - future II      &   36 &          83.3  & \textbf{100.0} &          58.3  & \textbf{100.0} &          86.1  &          83.3  &          86.1  & \textbf{100.0} &          72.2  &          63.9  &          50.0  &          77.8  & \textbf{100.0} &  \textbf{97.2} &          83.3  &          80.6  & 82.6 \\
Ditransitive - future II subjunctive II &   36 &          83.3  & \textbf{100.0} & \textbf{100.0} & \textbf{100.0} &  \textbf{94.4} &          83.3  &          83.3  & \textbf{100.0} &          80.6  & \textbf{100.0} & \textbf{100.0} &  \textbf{97.2} & \textbf{100.0} &          91.7  &          83.3  &          77.8  & 92.2 \\
Ditransitive - perfect        &   36 &          86.1  &  \textbf{97.2} & \textbf{100.0} & \textbf{100.0} & \textbf{100.0} &          83.3  & \textbf{100.0} & \textbf{100.0} &          88.9  &  \textbf{94.4} &  \textbf{97.2} & \textbf{100.0} & \textbf{100.0} & \textbf{100.0} &          83.3  & \textbf{100.0} & 95.7 \\
Ditransitive - pluperfect     &   35 &          48.6  & \textbf{100.0} &          51.4  &          60.0  &          82.9  &          57.1  &          91.4  &          45.7  &          25.7  &          25.7  &          20.0  &          51.4  &          91.4  &          80.0  &          77.1  &          74.3  & 61.4 \\
Ditransitive - pluperfect subjunctive II &   36 &          83.3  & \textbf{100.0} &  \textbf{97.2} &  \textbf{94.4} & \textbf{100.0} &          86.1  & \textbf{100.0} &          77.8  &          88.9  & \textbf{100.0} &  \textbf{94.4} &  \textbf{94.4} & \textbf{100.0} &          86.1  &          83.3  &  \textbf{94.4} & 92.5 \\
Ditransitive - present        &   36 &  \textbf{94.4} & \textbf{100.0} &  \textbf{97.2} &          86.1  &  \textbf{97.2} &          86.1  & \textbf{100.0} & \textbf{100.0} &          61.1  &          88.9  &          72.2  &          63.9  &  \textbf{97.2} &          66.7  &          77.8  & \textbf{100.0} & 86.8 \\
Ditransitive - preterite      &   35 &          80.0  &  \textbf{94.3} &          71.4  &          68.6  &          65.7  &          71.4  &          68.6  &  \textbf{85.7} &          77.1  &          60.0  &  \textbf{85.7} &          62.9  &  \textbf{85.7} &          74.3  &          65.7  &          62.9  & 73.8 \\
Ditransitive - preterite subjunctive II &   36 &          69.4  &          75.0  &          63.9  &          58.3  &          63.9  &          58.3  &          58.3  &          69.4  &          69.4  &          52.8  &          72.2  &          58.3  &          72.2  &          72.2  &          55.6  &          52.8  & 63.9 \\
Imperative                    &   20 &  \textbf{70.0} &  \textbf{85.0} &  \textbf{70.0} &  \textbf{65.0} &  \textbf{65.0} &  \textbf{70.0} &  \textbf{70.0} &  \textbf{85.0} &  \textbf{85.0} &  \textbf{60.0} &  \textbf{85.0} &          50.0  &  \textbf{80.0} &  \textbf{60.0} &  \textbf{65.0} &  \textbf{60.0} & 70.3 \\
Intransitive - future I       &   36 &  \textbf{97.2} &  \textbf{97.2} &  \textbf{97.2} &  \textbf{97.2} &          88.9  &  \textbf{97.2} & \textbf{100.0} & \textbf{100.0} & \textbf{100.0} & \textbf{100.0} & \textbf{100.0} & \textbf{100.0} &  \textbf{97.2} &  \textbf{97.2} &  \textbf{97.2} &  \textbf{94.4} & 97.6 \\
Intransitive - future I subjunctive II &   36 & \textbf{100.0} & \textbf{100.0} &          80.6  & \textbf{100.0} &          77.8  & \textbf{100.0} & \textbf{100.0} & \textbf{100.0} & \textbf{100.0} & \textbf{100.0} & \textbf{100.0} & \textbf{100.0} & \textbf{100.0} &          91.7  & \textbf{100.0} & \textbf{100.0} & 96.9 \\
Intransitive - future II      &   42 &          92.9  & \textbf{100.0} &          57.1  &          90.5  &          85.7  &          92.9  &  \textbf{97.6} &          81.0  &          88.1  &          78.6  & \textbf{100.0} &  \textbf{97.6} &  \textbf{95.2} &          85.7  &          85.7  &          85.7  & 88.4 \\
Intransitive - future II subjunctive II &   36 &  \textbf{97.2} & \textbf{100.0} &          83.3  &          75.0  &          88.9  &          91.7  & \textbf{100.0} & \textbf{100.0} &          77.8  & \textbf{100.0} & \textbf{100.0} & \textbf{100.0} &          88.9  &          83.3  &          75.0  &          83.3  & 90.3 \\
Intransitive - perfect        &   80 &  \textbf{97.5} &          92.5  &          90.0  &  \textbf{97.5} &          88.8  &          93.8  &  \textbf{98.8} &  \textbf{98.8} &  \textbf{97.5} & \textbf{100.0} & \textbf{100.0} &  \textbf{97.5} &  \textbf{98.8} &          85.0  &  \textbf{98.8} &          90.0  & 95.3 \\
Intransitive - pluperfect     &   36 &  \textbf{83.3} &          77.8  &          30.6  &          75.0  &          58.3  &          66.7  &  \textbf{94.4} &          47.2  &          19.4  &          44.4  &          16.7  &          55.6  &  \textbf{83.3} &          63.9  &          80.6  &          80.6  & 61.1 \\
Intransitive - pluperfect subjunctive II &   36 &  \textbf{97.2} & \textbf{100.0} &          77.8  & \textbf{100.0} &          91.7  &  \textbf{94.4} & \textbf{100.0} & \textbf{100.0} &          88.9  & \textbf{100.0} &  \textbf{97.2} &          86.1  &  \textbf{94.4} &          72.2  &          91.7  &          80.6  & 92.0 \\
Intransitive - present        &   36 & \textbf{100.0} & \textbf{100.0} &  \textbf{97.2} & \textbf{100.0} &          52.8  & \textbf{100.0} &  \textbf{97.2} &  \textbf{97.2} &  \textbf{97.2} & \textbf{100.0} &  \textbf{97.2} &  \textbf{94.4} & \textbf{100.0} &          91.7  & \textbf{100.0} &  \textbf{94.4} & 95.0 \\
Intransitive - preterite      &   65 &          80.0  &  \textbf{96.9} &          70.8  &          73.8  &          67.7  &          76.9  &          81.5  &          86.2  &          80.0  &          69.2  &  \textbf{96.9} &          75.4  &  \textbf{93.8} &          69.2  &          89.2  &          72.3  & 80.0 \\
Intransitive - preterite subjunctive II &   35 &  \textbf{65.7} &  \textbf{80.0} &  \textbf{60.0} &  \textbf{62.9} &          51.4  &  \textbf{68.6} &  \textbf{62.9} &  \textbf{65.7} &  \textbf{71.4} &          42.9  &  \textbf{71.4} &          54.3  &  \textbf{62.9} &          51.4  &  \textbf{71.4} &  \textbf{65.7} & 63.0 \\
Modal - future I              &  180 &          76.1  &  \textbf{77.2} &          75.6  &          71.1  &          61.1  &  \textbf{84.4} &  \textbf{80.0} &          74.4  &          65.0  &  \textbf{79.4} &  \textbf{78.3} &  \textbf{80.0} &  \textbf{78.3} &          73.3  &  \textbf{79.4} &  \textbf{80.0} & 75.9 \\
Modal - future I subjunctive II &  173 &          74.6  &          71.7  &          76.9  &          71.7  &          38.7  &  \textbf{81.5} &  \textbf{82.7} &          71.1  &          65.3  &          79.2  &          72.3  &          72.8  &  \textbf{82.7} &          61.3  &          77.5  &  \textbf{87.9} & 73.0 \\
Modal - perfect               &  168 &          88.7  &          73.2  &          72.6  &          83.3  &          34.5  &          83.9  &          62.5  &          69.0  &          73.2  &          42.3  &          91.7  &          85.7  &  \textbf{98.8} &          78.0  &          79.2  &          66.1  & 73.9 \\
Modal - pluperfect            &  179 &          20.1  &          29.1  &          11.2  &          40.2  &           7.3  &          22.9  &  \textbf{76.5} &          30.7  &           1.7  &           7.3  &           2.2  &          34.1  &          49.7  &          17.9  &          46.4  &          58.1  & 28.5 \\
Modal - pluperfect subjunctive II &  178 &  \textbf{57.3} &  \textbf{52.8} &  \textbf{55.6} &  \textbf{59.6} &          41.0  &  \textbf{59.6} &  \textbf{59.6} &  \textbf{52.2} &          42.7  &  \textbf{52.2} &  \textbf{49.4} &  \textbf{60.7} &  \textbf{56.2} &  \textbf{52.8} &  \textbf{59.0} &  \textbf{59.6} & 54.4 \\
Modal - present               &  179 &          90.5  &  \textbf{94.4} &          92.2  &  \textbf{93.3} &          48.6  &          86.6  &  \textbf{94.4} &  \textbf{96.6} &          59.8  &  \textbf{94.4} &          77.7  &          88.8  &  \textbf{95.0} &          85.5  &  \textbf{92.7} &  \textbf{96.1} & 86.7 \\
Modal - preterite             &  179 &          95.5  &  \textbf{97.2} &          86.6  &          96.6  &          52.0  &          89.4  &          93.9  &          95.0  &          89.4  &          95.0  &  \textbf{99.4} &          89.4  &  \textbf{99.4} &          86.0  &  \textbf{99.4} &          81.6  & 90.4 \\
Modal - preterite subjunctive II &  173 &  \textbf{75.7} &  \textbf{76.3} &  \textbf{72.8} &  \textbf{73.4} &          48.6  &  \textbf{73.4} &  \textbf{78.6} &  \textbf{72.8} &  \textbf{71.7} &  \textbf{77.5} &  \textbf{74.0} &  \textbf{74.0} &  \textbf{80.3} &          64.7  &  \textbf{71.7} &  \textbf{76.9} & 72.7 \\
Modal negated - future I      &  177 &  \textbf{76.3} &  \textbf{78.0} &  \textbf{75.7} &  \textbf{75.1} &          45.8  &  \textbf{81.4} &  \textbf{80.2} &          70.1  &          69.5  &  \textbf{80.2} &          70.1  &  \textbf{80.2} &  \textbf{78.5} &  \textbf{75.7} &  \textbf{79.7} &  \textbf{81.4} & 74.9 \\
Modal negated - future I subjunctive II &  175 &          78.3  &          71.4  &          76.6  &          77.7  &          60.6  &          83.4  &          84.0  &          69.7  &          67.4  &          81.7  &          72.0  &          78.9  &          81.7  &          69.7  &          81.7  &  \textbf{90.9} & 76.6 \\
Modal negated - perfect       &  175 &          93.1  &          73.1  &          80.6  &          92.6  &          65.1  &          83.4  &          91.4  &          69.1  &          68.0  &          77.1  &          70.9  &          86.9  &  \textbf{97.1} &          76.6  &          79.4  &          89.7  & 80.9 \\
Modal negated - pluperfect    &  173 &          10.4  &          13.9  &           0.0  &          34.7  &           8.7  &           6.4  &  \textbf{97.1} &          16.8  &           0.0  &          20.8  &           0.0  &          15.6  &          46.2  &           9.2  &          16.2  &          80.3  & 23.5 \\
Modal negated - pluperfect subjunctive II &  170 &          51.2  &          60.0  &          32.9  &  \textbf{64.1} &          51.2  &  \textbf{63.5} &  \textbf{68.8} &          33.5  &          38.2  &          43.5  &          50.0  &  \textbf{64.1} &  \textbf{65.3} &          58.8  &  \textbf{68.8} &  \textbf{70.6} & 55.3 \\
Modal negated - present       &  177 &  \textbf{99.4} &          96.0  &          90.4  &  \textbf{97.7} &          71.2  &          96.6  &  \textbf{97.2} &          68.9  &          72.3  &          77.4  &          67.2  &          92.7  &          96.6  &          83.6  &  \textbf{98.9} &          96.0  & 87.6 \\
Modal negated - preterite     &  178 &          93.8  &          96.6  &          83.7  &          98.3  &          79.8  &          89.3  &          98.3  &          96.1  &          81.5  &          93.8  &          94.4  &          88.2  & \textbf{100.0} &          91.0  &  \textbf{99.4} &          83.1  & 91.7 \\
Modal negated - preterite subjunctive II &  171 &          66.7  &  \textbf{74.3} &          64.9  &  \textbf{69.6} &          67.3  &  \textbf{73.7} &  \textbf{76.6} &  \textbf{72.5} &          66.1  &  \textbf{75.4} &  \textbf{69.0} &  \textbf{70.2} &  \textbf{77.2} &  \textbf{73.7} &  \textbf{73.7} &  \textbf{77.8} & 71.8 \\
Progressive                   &   20 &  \textbf{65.0} &  \textbf{85.0} &  \textbf{60.0} &  \textbf{70.0} &  \textbf{80.0} &          45.0  &          50.0  &  \textbf{60.0} &  \textbf{60.0} &          55.0  &  \textbf{80.0} &          45.0  &          55.0  &  \textbf{65.0} &  \textbf{70.0} &  \textbf{60.0} & 62.8 \\
Reflexive - future I          &   32 &  \textbf{87.5} &  \textbf{93.8} &  \textbf{87.5} &  \textbf{90.6} &  \textbf{87.5} &  \textbf{87.5} &  \textbf{81.2} &  \textbf{93.8} &  \textbf{81.2} &          65.6  &  \textbf{90.6} &  \textbf{81.2} &  \textbf{81.2} &          75.0  &  \textbf{84.4} &  \textbf{90.6} & 85.0 \\
Reflexive - future I subjunctive II &   36 &          75.0  &  \textbf{88.9} &          72.2  &  \textbf{80.6} &  \textbf{80.6} &  \textbf{83.3} &          69.4  &  \textbf{91.7} &  \textbf{83.3} &  \textbf{80.6} &  \textbf{91.7} &  \textbf{77.8} &          66.7  &          72.2  &  \textbf{80.6} &          75.0  & 79.3 \\
Reflexive - future II         &   33 &          75.8  &  \textbf{84.8} &          33.3  &          78.8  &          66.7  &  \textbf{90.9} &          69.7  &  \textbf{87.9} &          48.5  &          33.3  &          81.8  &          57.6  &  \textbf{97.0} &          72.7  &          81.8  &  \textbf{90.9} & 72.0 \\
Reflexive - future II subjunctive II &   34 &  \textbf{82.4} &  \textbf{94.1} &          70.6  &          70.6  &          67.6  &  \textbf{85.3} &          67.6  &  \textbf{88.2} &          61.8  &          47.1  &  \textbf{88.2} &          70.6  &          76.5  &          73.5  &  \textbf{82.4} &          64.7  & 74.4 \\
Reflexive - perfect           &   32 &  \textbf{96.9} &  \textbf{90.6} &          68.8  &  \textbf{90.6} &  \textbf{84.4} &  \textbf{93.8} &          78.1  &  \textbf{93.8} &          68.8  &          68.8  &  \textbf{87.5} &          68.8  &  \textbf{87.5} &          81.2  &  \textbf{84.4} &  \textbf{96.9} & 83.8 \\
Reflexive - pluperfect        &   31 &          74.2  &          80.6  &          71.0  &          80.6  &          67.7  &  \textbf{96.8} &          74.2  &  \textbf{93.5} &          67.7  &          22.6  &          80.6  &          64.5  &  \textbf{83.9} &  \textbf{93.5} &          77.4  &  \textbf{90.3} & 76.2 \\
Reflexive - pluperfect subjunctive II &   34 &  \textbf{76.5} &  \textbf{79.4} &  \textbf{79.4} &  \textbf{76.5} &  \textbf{67.6} &  \textbf{79.4} &  \textbf{61.8} &  \textbf{76.5} &  \textbf{79.4} &          47.1  &  \textbf{79.4} &  \textbf{61.8} &  \textbf{70.6} &  \textbf{67.6} &  \textbf{79.4} &  \textbf{82.4} & 72.8 \\
Reflexive - present           &   35 &          80.0  &  \textbf{82.9} &          77.1  &          65.7  &          65.7  &  \textbf{91.4} &  \textbf{82.9} &  \textbf{94.3} &          54.3  &          57.1  &          80.0  &          57.1  &          68.6  &          71.4  &          80.0  &  \textbf{82.9} & 74.5 \\
Reflexive - preterite         &   32 &          75.0  &  \textbf{96.9} &          50.0  &          78.1  &          56.2  &          71.9  &          68.8  &  \textbf{84.4} &  \textbf{84.4} &          25.0  &  \textbf{87.5} &          46.9  &          75.0  &          75.0  &          81.2  &          68.8  & 70.3 \\
Reflexive - preterite subjunctive II &   34 &  \textbf{70.6} &  \textbf{88.2} &          47.1  &  \textbf{73.5} &          44.1  &          61.8  &          58.8  &  \textbf{76.5} &  \textbf{73.5} &          20.6  &  \textbf{82.4} &          47.1  &  \textbf{76.5} &  \textbf{73.5} &  \textbf{73.5} &          58.8  & 64.2 \\
Transitive - future I         &   41 &         100.0  &         100.0  &         100.0  &         100.0  &          97.6  &         100.0  &         100.0  &         100.0  &         100.0  &         100.0  &         100.0  &         100.0  &         100.0  &         100.0  &         100.0  &         100.0  & 99.8 \\
Transitive - future I subjunctive II &   36 &         100.0  &         100.0  &          97.2  &         100.0  &         100.0  &         100.0  &         100.0  &         100.0  &         100.0  &         100.0  &         100.0  &         100.0  &         100.0  &          97.2  &         100.0  &         100.0  & 99.7 \\
Transitive - future II        &   36 & \textbf{100.0} & \textbf{100.0} &          86.1  & \textbf{100.0} &  \textbf{94.4} & \textbf{100.0} & \textbf{100.0} & \textbf{100.0} & \textbf{100.0} &          91.7  & \textbf{100.0} & \textbf{100.0} & \textbf{100.0} &  \textbf{94.4} & \textbf{100.0} &  \textbf{97.2} & 97.7 \\
Transitive - future II subjunctive II &   36 & \textbf{100.0} & \textbf{100.0} & \textbf{100.0} &          83.3  &  \textbf{94.4} & \textbf{100.0} & \textbf{100.0} & \textbf{100.0} & \textbf{100.0} &  \textbf{94.4} & \textbf{100.0} & \textbf{100.0} & \textbf{100.0} &  \textbf{94.4} & \textbf{100.0} &          83.3  & 96.9 \\
Transitive - perfect          &   41 &  \textbf{95.1} & \textbf{100.0} & \textbf{100.0} & \textbf{100.0} & \textbf{100.0} &  \textbf{97.6} & \textbf{100.0} & \textbf{100.0} & \textbf{100.0} &          92.7  & \textbf{100.0} &          87.8  & \textbf{100.0} & \textbf{100.0} & \textbf{100.0} & \textbf{100.0} & 98.3 \\
Transitive - pluperfect       &   36 & \textbf{100.0} & \textbf{100.0} &          72.2  &          69.4  & \textbf{100.0} &          80.6  & \textbf{100.0} &          72.2  &          44.4  &          91.7  &          41.7  &          83.3  & \textbf{100.0} &          91.7  & \textbf{100.0} &          88.9  & 83.5 \\
Transitive - pluperfect subjunctive II &   36 &  \textbf{94.4} & \textbf{100.0} &  \textbf{97.2} &  \textbf{97.2} & \textbf{100.0} & \textbf{100.0} & \textbf{100.0} &  \textbf{94.4} &  \textbf{97.2} &  \textbf{97.2} & \textbf{100.0} &  \textbf{97.2} & \textbf{100.0} &          91.7  & \textbf{100.0} &  \textbf{97.2} & 97.7 \\
Transitive - present          &   48 & \textbf{100.0} & \textbf{100.0} & \textbf{100.0} & \textbf{100.0} &  \textbf{97.9} & \textbf{100.0} &  \textbf{97.9} & \textbf{100.0} &          93.8  &  \textbf{97.9} &  \textbf{95.8} &  \textbf{97.9} & \textbf{100.0} &          85.4  & \textbf{100.0} & \textbf{100.0} & 97.9 \\
Transitive - preterite        &   36 &          86.1  &  \textbf{97.2} &          80.6  &          80.6  &          69.4  &          77.8  &          72.2  &          83.3  &  \textbf{97.2} &          72.2  &  \textbf{97.2} &          80.6  & \textbf{100.0} &          83.3  &          86.1  &          72.2  & 83.5 \\
Transitive - preterite subjunctive II &   36 &          47.2  &  \textbf{83.3} &          58.3  &          61.1  &          47.2  &  \textbf{66.7} &          55.6  &          58.3  &  \textbf{75.0} &          30.6  &  \textbf{63.9} &          52.8  &  \textbf{63.9} &          44.4  &  \textbf{75.0} &          58.3  & 58.9 \\
\rowcolor{Gray}
Verb valency                  &   86 &  \textbf{72.1} &  \textbf{79.1} &  \textbf{68.6} &  \textbf{67.4} &  \textbf{70.9} &  \textbf{66.3} &  \textbf{67.4} &  \textbf{68.6} &  \textbf{67.4} &          55.8  &  \textbf{66.3} &          54.7  &  \textbf{72.1} &          62.8  &  \textbf{68.6} &          60.5  & 66.8 \\
Case government               &   27 &          77.8  &  \textbf{96.3} &  \textbf{81.5} &          74.1  &  \textbf{81.5} &          74.1  &          70.4  &          74.1  &          77.8  &          63.0  &          70.4  &          55.6  &  \textbf{81.5} &          70.4  &          70.4  &          63.0  & 73.8 \\
Mediopassive voice            &   20 &          85.0  &          85.0  &          70.0  &          75.0  &          80.0  &          75.0  &          80.0  &          75.0  &          70.0  &          60.0  &          80.0  &          60.0  &          80.0  &          65.0  &          80.0  &          70.0  & 74.4 \\
Passive voice                 &   20 & \textbf{100.0} & \textbf{100.0} &  \textbf{95.0} & \textbf{100.0} & \textbf{100.0} &  \textbf{95.0} & \textbf{100.0} & \textbf{100.0} &  \textbf{95.0} &          80.0  &  \textbf{90.0} &  \textbf{95.0} & \textbf{100.0} &  \textbf{95.0} & \textbf{100.0} &  \textbf{95.0} & 96.2 \\
Resultative predicates        &   19 &          21.1  &          26.3  &          21.1  &          15.8  &          15.8  &          15.8  &          15.8  &          21.1  &          21.1  &          15.8  &          21.1  &           5.3  &          21.1  &          15.8  &          21.1  &          10.5  & 17.8 \\
\midrule
average (items)               & 5393 &          78.0  &          80.9  &     
71.6  &          79.2  &          64.3  &          77.7  &  \textbf{82.8} &          75.5  &          67.5  &          68.4  &          74.1  &          74.4  &  \textbf{83.6} &          72.3  &          79.2  &          80.2  & 75.6 \\
\bottomrule
\\
\caption{Accuracies (\%) of successful translations for 16 systems and
107 phenomena organized in 14 categories. Boldface indicates the significantly
    best systems in each row.}
    \label{tab:phenomena}
\end{longtable}

} 

\label{sec:appendix}

\end{landscape}
\end{document}